\documentclass[10pt,journal,compsoc]{IEEEtran}

\ifCLASSOPTIONcompsoc
  \usepackage[nocompress]{cite}
\else
  \usepackage{cite}
\fi
\ifCLASSINFOpdf
\else
\fi
\usepackage{amsmath,amssymb,amsfonts}
\usepackage{algorithmic}
\usepackage[ruled,vlined,linesnumbered]{algorithm2e}
\usepackage{algorithmic,algorithm2e,float}
\SetKwInput{KwInput}{Input}                
\SetKwInput{KwOutput}{Output}              
\SetKwInput{KwRepeat}{Repeat}
\SetKwInput{KwUntil}{Until}
\SetKwInput{KwReturn}{Return}
\usepackage{amsfonts}
\usepackage{graphicx}
\usepackage{textcomp}
\usepackage{xcolor}
\usepackage{tabularx}
\usepackage{array}
\usepackage{booktabs}
\usepackage{threeparttable}  
\usepackage{multirow}
\usepackage{bm}
\usepackage{subfigure}

\usepackage{amssymb}
\usepackage{pifont}
\newcommand\com[1]{{\color{blue}#1}}
\newcolumntype{b}{X}
\newcolumntype{s}{>{\hsize=.4\hsize}X}
\newcolumntype{m}{>{\hsize=.6\hsize}X}
\newcolumntype{l}{>{\hsize=1.2\hsize}X}

\begin{document}
%
\title{TSception: Capturing Temporal Dynamics and Spatial Asymmetry from EEG for Emotion Recognition}
%
%
%
%
\author{Yi~Ding,~\IEEEmembership{Student Member,~IEEE,}
        Neethu~Robinson,~\IEEEmembership{Member,~IEEE,}
        Su~Zhang,
        Qiuhao~Zeng,
        and~Cuntai~Guan,~\IEEEmembership{Fellow,~IEEE}
\IEEEcompsocitemizethanks{\IEEEcompsocthanksitem Yi Ding, Neethu Robinson, Su Zhang, Qiuhao Zeng, and Cuntai Guan are with the School of Computer Science and Engineering, Nanyang Technological University, 50 Nanyang Avenue, Singapore, 639798.\protect\\
E-mail: (ding.yi, nrobinson, su012, qiuhao.zeng, ctguan)@ntu.edu.sg.}
\thanks{Cuntai Guan is the Corresponding Author.}}

\markboth{IEEE Transactions on Affective Computing}%
{Shell \MakeLowercase{\textit{et al.}}: Bare Advanced Demo of IEEEtran.cls for IEEE Computer Society Journals}

\IEEEtitleabstractindextext{%
\begin{abstract}
The high temporal resolution and the asymmetric spatial activations are essential attributes of electroencephalogram (EEG) underlying emotional processes in the brain. To learn the temporal dynamics and spatial asymmetry of EEG towards accurate and generalized emotion recognition, we propose TSception, a multi-scale convolutional neural network that can classify emotions from EEG. TSception consists of dynamic temporal, asymmetric spatial, and high-level fusion layers, which learn discriminative representations in the time and channel dimensions simultaneously. The dynamic temporal layer consists of multi-scale 1D convolutional kernels whose lengths are related to the sampling rate of EEG, which learns the dynamic temporal and frequency representations of EEG. The asymmetric spatial layer takes advantage of the asymmetric EEG patterns for emotion, learning the discriminative global and hemisphere representations. The learned spatial representations will be fused by a high-level fusion layer. Using more generalized cross-validation settings, the proposed method is evaluated on two publicly available datasets DEAP and MAHNOB-HCI. The performance of the proposed network is compared with prior reported methods such as SVM, KNN, FBFgMDM, FBTSC, Unsupervised learning, DeepConvNet, ShallowConvNet, and EEGNet. TSception achieves higher classification accuracies and F1 scores than other methods in most of the experiments. The codes are available at:\textit{https://github.com/yi-ding-cs/TSception}
\end{abstract}

}

\maketitle

\IEEEdisplaynontitleabstractindextext

%
\IEEEpeerreviewmaketitle

\ifCLASSOPTIONcompsoc
\IEEEraisesectionheading{\section{Introduction}\label{sec:introduction}}
\else
\section{Introduction}
\label{sec:introduction}
\fi

%
%
%
%
\IEEEPARstart{E}{motions} are fundamental factors in human beings' daily life \cite{7946165}, affecting decision-making, perception, human interaction, and human intelligence \cite{Dolan1191}. Emotion recognition plays an important role in Cognitive Behavioural Therapy (CBT) \cite{doi:10.1002/da.22728}, Emotion Regulation Therapy (ERT)/Emotion-Focused Therapy (EFT) \cite{FRESCO2013282}\cite{doi:10.1002/cpp.388}\cite{lane_ryan_nadel_greenberg_2015}, and the evaluation of medical treatment \cite{doi:10.1176/appi.ajp.162.9.1746} for emotion-related mental disorders, such as Generalized Anxiety Disorder (GAD) \cite{GOODWIN2017107}, and Depression \cite{Duman2016}. With the potential applications in CBT and EFT, enabling Artificial Intelligence (AI) to identify human emotions has captured more and more interest from researchers recently \cite{7946165}. 

Electroencephalography (EEG) is one of the widely used brain imaging technologies, which measures human brain activity directly. Several electrodes are placed on the surface of the human head to collect EEG signals. EEG has high temporal resolution so that it can capture varying brain states at the sub-second level. A Brain-Computer Interface (BCI) system can identify human emotions through EEG, with the help of machine learning and signal processing techniques \cite{10.1371/journal.pone.0213516}. 

Recently, using EEG-BCI for emotion recognition has gained popularity among researchers \cite{7946165}\cite{Craik_2019}. Atkinson \textit{et al}. \cite{ATKINSON201635} improved the SVM classifier accuracy for emotion detection by selecting features efficiently, with the accuracy being 73.14\%. Zheng \textit{et al} \cite{7938737} used a discriminative graph regularized extreme learning machine to investigate stable patterns over time from the differential entropy (DE) features of emotional EEG. Li \textit{et al.} \cite{8634938} utilized phase-locking value to construct emotion-related brain networks with multiple feature fusion to detect emotions from EEG. Recently, deep learning-based methods have shown promising results in the BCI domain, such as motor imagery classification \cite{doi:10.1002/hbm.23730}\cite{8897723}\cite{Tabar_2016}\cite{Lawhern_2018}\cite{8310961}, emotion recognition \cite{7883875}\cite{Li2018}\cite{10.3389/fnins.2018.00162}\cite{8567966}\cite{Zhang_Cui_Xu_Zheng_Yang_2020}\cite{8275511}, and mental-task classification \cite{Fahimi_2019}\cite{JIAO2018582}\cite{8607897}. Yang \textit{et al.} \cite{7883875} designed a hierarchical network structure to perform emotion classification, proposing sub-network nodes to enhance the performance. Li \textit{et al.} \cite{Li2018} constructed EEG into 2D images and proposed a Hierarchical Convolutional Neural Networks (HCNN) to extract the spatial patterns of the EEG. Li \textit{et al.} \cite{10.3389/fnins.2018.00162} applied 18 kinds of linear and non-linear features to solve the cross-subject emotion recognition problems, achieving 59.06\% and 83.33\% on two public datasets. Zhang \textit{et al.} \cite{8275511} utilized recurrent neural networks (RNN) to learn the temporal-spatial information from the DE features of EEG for emotion recognition. Although many machine learning methods have been proposed for emotion recognition, most of them highly rely on hand-crafted features. 

With the ability to learn from EEG directly, the convolutional neural networks (CNN) have shown promising results in BCI \cite{Lawhern_2018}\cite{8914184}\cite{9175874}. Schirrmeister \textit{et al.} \cite{doi:10.1002/hbm.23730} proposed deep and shallow convolutional neural networks, named DeepConvNet and ShallowConvNet, to process EEG data, combining the feature extraction and classification using a two-stage spatial and temporal input convolution layer. Recently, Lawhern \textit{et al.} \cite{Lawhern_2018} proposed EEGNet,  which extracts spatial information by the depth-wise convolution kernel whose size is $(n,1)$. The global spatial dependency can be learned by letting $n$ be the number of channels. All of those networks apply single-scale 1D convolutional kernels along the time and channel dimension to extract temporal and spatial information from EEG. 

In order to effectively learn temporal-spatial information from EEG for emotion recognition, several neuro-physiological signatures should be considered. For temporal dimension, EEG signal contains abundant brain activity information in different frequency bands \cite{4634130}. Due to the non-stationary and dynamic nature of EEG, we hypothesize that a single-sized temporal kernel cannot effectively capture the neural processing underlying emotions that occurs at different time scales and duration. 
For spatial dimension, especially for emotional processes in the brain, the right and left hemispheres have asymmetric responses to emotions \cite{doi:10.1111/psyp.13028}. Hence, we hold the hypothesis that a global spatial kernel has less ability to effectively extract the distinct asymmetric EEG pattern during emotional processes. 

To address the above issues, in this paper, we propose TSception, a multi-scale temporal-spatial convolutional neural network to capture temporal dynamics and spatial asymmetry from EEG to classify emotional states. Different from the methods using manually extracted features \cite{Li2018}\cite{10.3389/fnins.2018.00162}\cite{8275511}\cite{5871728}\cite{9206731}, EEG signals are fed into TSception directly, which makes it an end-to-end deep learning method that needs less domain knowledge about the features. A dynamic temporal layer with different scaled convolutional kernels is proposed to learn richer time-frequency representations from EEG instead of using single-sized temporal CNN kernels \cite{doi:10.1002/hbm.23730}\cite{Lawhern_2018}. This layer is inspired by the inception block of GoogleNet \cite{Szegedy_2015_CVPR}. Besides the global kernel utilized in \cite{doi:10.1002/hbm.23730}\cite{Lawhern_2018}, we take the brain emotional asymmetry into the kernel design. A hemisphere kernel whose length equals the number of EEG channels located on the right/left hemisphere is proposed to extract the hemisphere asymmetric pattern. The effectiveness of multi-scale convolutional neural networks is preliminarily explored in our previous work \cite{9206750}. We further propose a high-level fusion layer after asymmetric spatial layer to learn from combined hemisphere-global representations to distinct emotion-class specific information as well as make the network more compact for online usage in the future. 

Emotion classification experiments on two publicly available benchmark datasets, a Database for Emotion Analysis using Physiological signals (DEAP) \cite{5871728}, and a multimodal database for affect recognition and implicit tagging (MAHNOB-HCI) \cite{soleymani2011multimodal} were conducted to evaluate the performance of TSception. The generalized cross-validation settings are utilized to avoid potential data leakage and biased evaluation. TSception is compared with several deep and non-deep state-of-the-art methods in the BCI domain, namely SVM \cite{5871728}, KNN \cite{9206731}, DeepConvNet \cite{doi:10.1002/hbm.23730}, ShallowConvNet \cite{doi:10.1002/hbm.23730}, EEGNet \cite{Lawhern_2018}, Unsupervised learning \cite{LIANG2019257}, FBFgMDM \cite{9141493}, and FBTSC \cite{9141493}. In most of the experiments, the performance of TSception in terms of accuracy and F1 score is higher than the other methods while having a relatively lesser number of network parameters. After statistical analysis, extensive ablation studies are conducted to analyze the contribution of each module in TSception. The saliency map method \cite{simonyan2013deep} is utilized to get the most informative part of the EEG data identified by the network. The maps show that the network mainly learns from frontal, temporal, and parietal areas. Frontal, parietal and temporal are commonly known as the functional brain areas related to the emotional processes in the brain \cite{7946165}.

The major contributions of this work can be summarised as: 
\begin{itemize}
\item We propose TSception, a novel multi-scale temporal-spatial convolutional neural network, for EEG emotion recognition tasks. Several neuro-physiological signatures are involved in the network design. The proposed multi-scale temporal/spatial convolution kernels can capture temporal dynamics and spatial asymmetry from EEG to classify emotions. A high-level fusion layer is proposed to further learn from hemisphere-global representations and to make the network more compact, which can benefit the online usage of TSception in the future. 
\item Extensive ablation studies and interpretability experiments are conducted to understand the importance of each module in TSception and what it learns using saliency maps.
\end{itemize}

The PyTorch implementation of TSception is available at \textit{https://github.com/yi-ding-cs/TSception}

The remainder of this article is organized as follows. A summary of related works is introduced in Section II. In Section III the details of TSception are introduced. Section IV describes the datasets and experiment settings. The result and analysis are given in Section V, Finally, we discuss the significance of our results in Section VI.

\begin{figure*}[htp]
    \centering
    \includegraphics[width= 18cm]{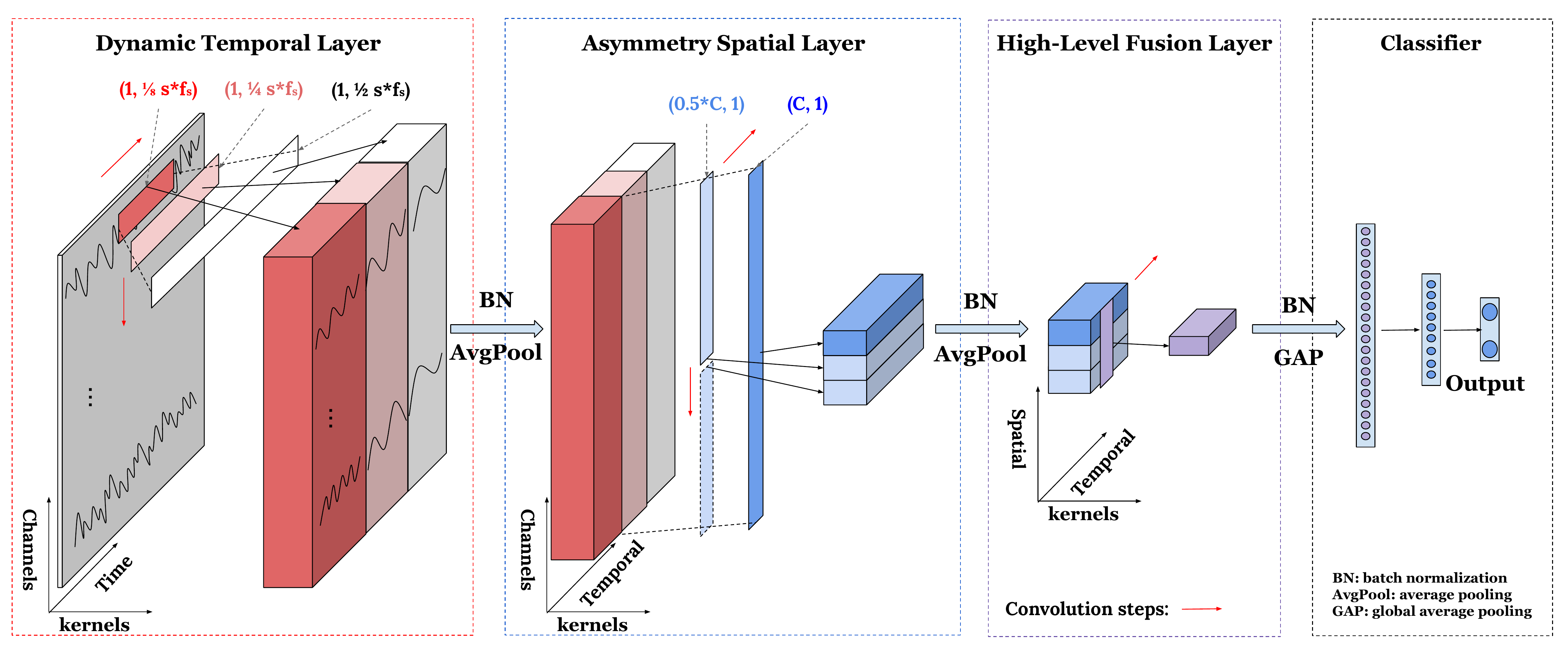}
    \caption{Structure of TSception. In the figure, $f_{s}$ is the sampling rate of the EEG signals, $C$ is the number of channels, BN stands for batch normalization, AP is the average pooling operation, and GAP represents global average pooling. TSception has four main parts: the dynamic temporal layer, the asymmetric spatial layer, the high-level fusion layer, and the classifier. The dynamic temporal layer will first learn the dynamic temporal/frequency representations from EEG data channel by channel. After getting the learned representations for each channel, the asymmetric spatial layer will be applied to learn the global spatial representations and the emotional asymmetry pattern using different scale convolutional kernels. To fuse the information from hemisphere and global representations, a high-level fusion layer is utilized. Finally, the fused representation will be passed to the fully connected layers with the softmax as the activation function.}
    \label{fig:TS-II}
\end{figure*}
\section{Multi-scale Convolutional Neural Networks}

The detailed instruction of the proposed TSception, a multi-scale convolutional neural network, is presented in this section. EEG data can be treated as 2D time series, whose dimensions are channels (EEG electrodes) and time respectively. The time dimension reflects the brain activity changes from time to time. The spatial dimension can show the brain activation patterns across different functional areas due to the different locations of the electrodes on the brain. EEG signals contain abundant information in different frequency bands \cite{4634130}. TSception is proposed to identify the most distinct time-frequency-channel specific EEG features corresponding to the emotional states of the user. TSception incorporates specially designed network modules namely, dynamic temporal layer, asymmetric spatial layer, and high-level fusion layer. To extract more discriminative time-frequency representations, multi-scale 1D convolutional kernels are utilized in the dynamic temporal layer to enrich the learned time-frequency representations. As for the asymmetric spatial layer, it takes the advantage of neuroscience findings \cite{doi:10.1111/psyp.13028} which indicate the brain activities in right and left hemispheres are not symmetrically related to emotions. A hemisphere kernel is proposed to learn the asymmetric representations between two hemispheres. A high-level fusion layer is further proposed to learn from the learned representations of both the hemisphere and global kernels and make the network more compact for real-time usage. The network structure of TSception is shown in Fig.~\ref{fig:TS-II}. A detailed description of the temporal, spatial, and high-level fusion layers will be discussed in this section.

\subsection{Dynamic Temporal Layer}
The dynamic temporal layer consists of multi-scale 1D temporal kernels (T kernels). In order to enable the neural network to learn dynamic temporal representations, we set the length of the temporal kernels as the specific ratios of sampling rate $f_S$ of EEG. These ratios are defined as $ \alpha^{i} \in \mathbb{R}$, where $i$ is the level of the dynamic temporal layer. $i$ will vary from 1 to $L$, if the dynamic temporal layer has $L$ levels. Hence $s_{T}^{i}$, the size of T kernels in $i$-th level, can be defined as:
\begin{equation}\label{eq:size_t}
    s_{T}^{i} = \left ( 1, \alpha^{i} \cdot f_{S}\right), i \in [1, 2, 3]
\end{equation}

From the frequency perspective, the length of the T kernel is set as half the sampling rate in EEGNet, allowing for capturing frequency information at 2 Hz and above \cite{Lawhern_2018}. Activations related to emotions are observed in Alpha (8-12 Hz), Beta (12-30 Hz), and Gamma ($>$30 Hz) bands \cite{7946165}. In this work, we expand the temporal receptive-field, letting $L=3, i=1$ to 3, and $\alpha = 0.5$, the ratio coefficients will become [0.5, 0.25, 0.125], learning diversified frequency representations. We hypothesize that the multi-scale temporal kernels can enrich the learned dynamic frequency representations from EEG, providing more emotion-related information. From the time perspective, multi-scale T kernels can capture long short-term temporal patterns, and learn more diverse representations. The higher level T kernel has a smaller ratio coefficient, which gives a shorter convolutional kernel length and vice versa. The long temporal kernel can learn long-term temporal and low-frequency diverse representations. The short kernel extracts short-term temporal and high-frequency representations. Let $\textit{\textbf{X}}$ denote EEG input samples. $\textit{\textbf{X}} = \left [ \textit{\textbf{X}}^{0} ,\textit{\textbf{X}}^{1}, ... , \textit{\textbf{X}}^{n} \right ], \textit{\textbf{X}}^{n} \in \mathbb{R}^{c \times l}$, where $n$ is the number of EEG samples, $c$ is the number of channels, $l$ is the length of each sample. 
The dynamic temporal representations can be generated by parallelly applying the multi-scale temporal kernels on the input EEG samples. After $LeakyReLU(\cdot)$ activation function, the feature map is further down-sampled by average pooling (AP). The reason for using average pooling is to reduce the effect of the noise as well as the feature dimension since EEG signals are of high dimensions with a low signal-noise ratio. Let $\textit{\textbf{Z}}_{temporal}^{i}$ denote the output of the $i$-th level temporal kernel, $\textit{\textbf{Z}}_{temporal}^{i} \in \mathbb{R}^{n \times t \times c \times f_{i} }$, where $n$ is the number of samples, $t$ is the number of each level's T kernel, $c$ is the number of channels, and $f_{i}$ is the length of the feature after $i$-th level convolution operation. $\textit{\textbf{Z}}_{temporal}^{i}$ is defined as:
 \begin{equation}\label{eq:out_T}
     \textit{\textbf{Z}}_{temporal}^{i} = AP(\Phi_{L-ReLU}(Conv1D(\textit{\textbf{X}}, s_{T}^{i})))
 \end{equation}
 where $s_{T}^{i}$ is the T kernel size, $\textit{\textbf{X}}$ is the input EEG sample array, $Conv1D(\cdot)$ is the 1D convolution operation with the kernel size being $s_{T}^{i}$, step being (1,1), and $\Phi_{L-ReLU}(\cdot)$ is the LeakyReLU$(\cdot)$ activation function. 
 
 The output of each level's T kernel will be concatenated along the feature dimension. In order to reduce the internal covariate shift problems in neural networks, we added batch normalization \cite{ioffe2015batch} after the dynamic temporal layer. Hence the final output of the dynamic temporal layer, $\textbf{Z}_{T}$, $\textbf{Z}_{T} \in \mathbb{R}^{n \times t \times c \times \sum f_{i}}$, is defined as:
 \begin{equation}\label{eq:final_out_T}
     \textit{\textbf{Z}}_{T} = f_{bn}([\textit{\textbf{Z}}_{temporal}^{1},...,\textit{\textbf{Z}}_{temporal}^{i}]), i\in[1, 2, 3]
 \end{equation}
 where $f_{bn}$ is the batch normalization operation, and $[\cdot ]$ stands for concatenation operation along the feature (f) dimension.

\subsection{Asymmetric Spatial Layer}
The asymmetric spatial layer has multi-scale 1D convolutional kernels whose sizes are related to the location of the EEG channels. There are two types of spatial kernels: global kernel and hemisphere kernel. 

The global kernel has a size of $(c,1)$, where $c$ is the number of channels. Since the length of the kernel is the same as the channel dimension of the input EEG segment, it can learn the global spatial information. 

In this work, we further combine the frontal area of brain emotional asymmetry \cite{CRAIG2005566} into the kernel design. The hemisphere kernel is used to extract the relations between the left and right hemispheres by sharing the convolutional kernels. The size of the hemisphere kernel is $(0.5 \cdot c, 1)$, and the step is $(0.5 \cdot c, 1)$, where $c$ is the total number of channels. The hemisphere kernel is shared by two hemispheres without overlapping so that the asymmetric pattern can be extracted. The size of the spatial kernel $s_{S}^{j}$ can be defined as:
\begin{equation}\label{eq:size_S}
    s_{S}^{j} = \left (\delta^{j} \cdot c , 1\right), j \in [0, 1]
\end{equation}
where $\delta = 0.5$ is the coefficient to control the ratio between the spatial kernel length and the total number of channels.

Let $\textit{\textbf{Z}}_{spatial}^{j}$ denote the output of the $j$-th type spatial kernel, $\textit{\textbf{Z}}_{spatial}^{j} \in \mathbb{R}^{n \times s \times c_{j} \times f }$, where $n$ is the number of samples, $s$ is the number of each type S kernel, $c_{j}$ is the number of channels after $j$-th spatial convolution, and $f$ is the length of the feature after each spatial convolution operation. $\textit{\textbf{Z}}_{spatial}^{j}$ is defined as:
 \begin{equation}\label{eq:out_S}
     \textit{\textbf{Z}}_{spatial}^{j} = AP(\Phi_{L-ReLU}(Conv1D(\textit{\textbf{Z}}_{T}, s_{S}^{j})))
 \end{equation}
 where $s_{S}^{j}$ is the S kernel size, $\textit{\textbf{Z}}_{T}$ is the output of dynamic temporal layer, $Conv1D(\cdot)$ is the 1D convolution operation with the kernel size being $s_{S}^{j}$, the step being (1,1) for the global kernels and $(0.5 \cdot c, 1)$ for the hemisphere kernels, and $\Phi_{L-ReLU(\cdot)}$ is the LeakyReLU$(\cdot)$ activation function. 
\begin{figure}[htp]
    \centering
    \includegraphics[width= 6cm]{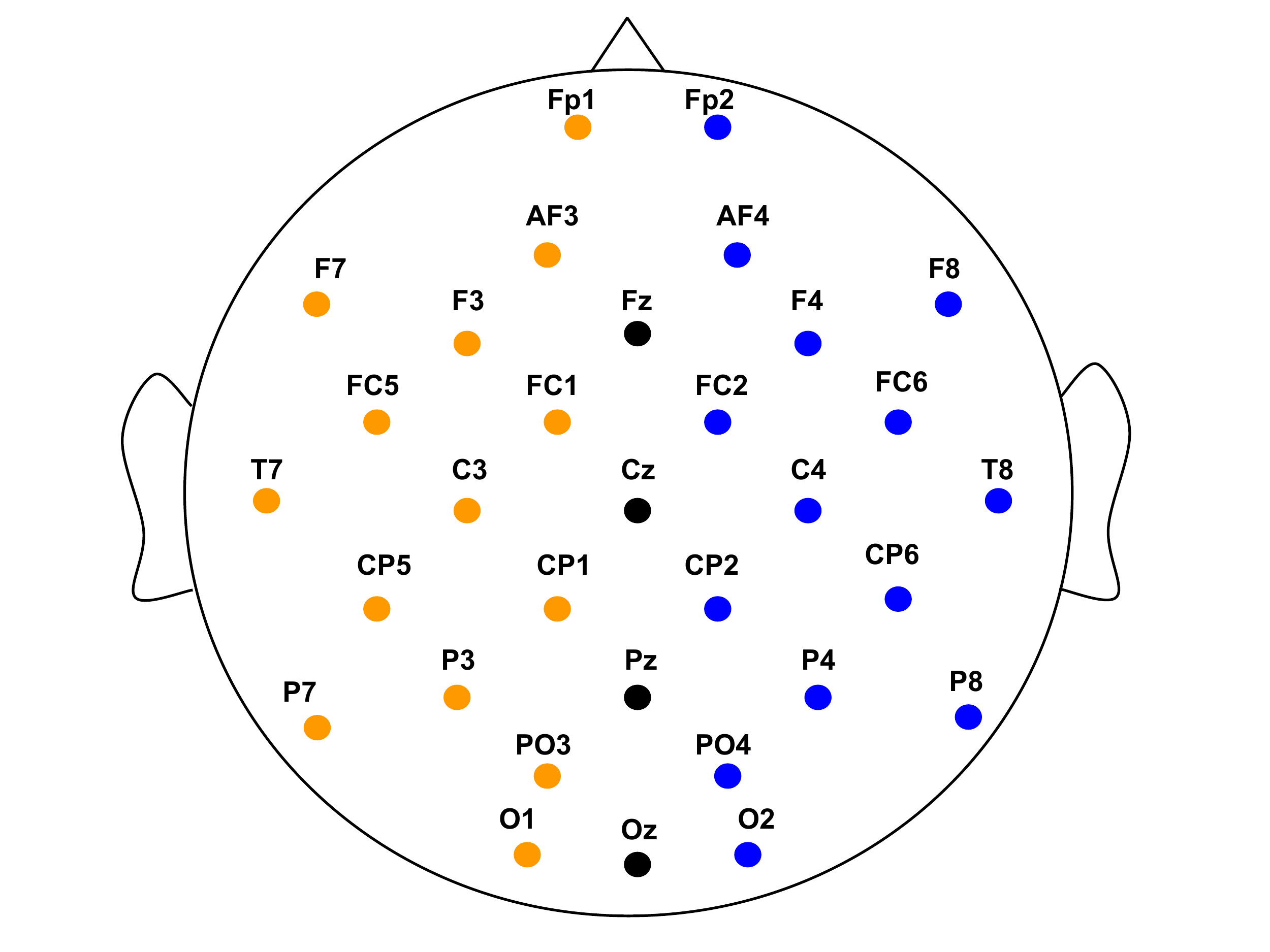}
    \caption{The location map of 32 channels cap. The electrodes can be divided into 3 groups: electrodes on the left hemisphere (in orange), electrodes on the right hemisphere (in blue), and the electrodes on the central line (in black). For the electrodes located on the central line of the head, Fz, Cz, Pz, and Oz, which can not be paired on the left and right hemispheres, we further removed them to let TSception learn the asymmetric pattern of left and right hemispheres better.}
    \label{fig:32_channel}
\end{figure}

In order to apply hemisphere kernels, the sequence of channels in the input EEG samples should be arranged in a particular way. The order of the channels should be $[ channel_{left}, channel_{right}]$, where the $channel_{left}$ are the channels located in the left hemisphere, the $channel_{right}$ are the ones on the right hemisphere. The order for channels on each hemisphere should also be rearranged to make each kernel weight shared between pairs of symmetrically located electrodes on two hemispheres because the step of the hemisphere kernel is also $(0.5 \cdot c, 1$). Fig.~\ref{fig:32_channel} shows the electrode locations of DEAP dataset. The final output of the asymmetric spatial layer, $\textit{\textbf{Z}}_{S}$, $\textit{\textbf{Z}}_{S} \in \mathbb{R}^{n \times s \times \sum c_{j} \times f}$ is defined as:
 \begin{equation}\label{eq:final_out_S}
     \textit{\textbf{Z}}_{S} = f_{bn}([\textit{\textbf{Z}}_{spatial}^{0},...,\textit{\textbf{Z}}_{spatial}^{j}]), j\in[0, 1]
 \end{equation}
 where $f_{bn}$ is the batch normalization operation, and $[\cdot ]$ stands for concatenation operation along the channel ($c$) dimension. The output of hemisphere kernel have a length of two in the spatial dimension, which refers to two hemispheres respectively. The output of global kernel is only a vector whose length in the channel dimension is one. After concatenation, the channel dimension is $\sum c_{j}=3$.

\subsection{High-level Fusion Layer}
In order to learn high-level spatial representations by fusing the learned information from global and hemispheres, a high-level fusion layer is further proposed. Given the output of asymmetric spatial layer, $\textit{\textbf{Z}}_{S} \in \mathbb{R}^{n \times s \times 3 \times f}$, a 1D convolutional layer whose kernel size is $(3, 1)$ is utilized to fuse the information along the spatial dimension. After LeakyReLU($\cdot$), average pooling, and batch normalization, a global average pooling layer (GAP) is added to overcome over-fitting and reduce the model size. The final learned global-hemisphere fusion representations will be generated by: 
\begin{equation}\label{eq:final_out_G_H}
     \textit{\textbf{Z}}_{fusion} = GAP(f_{bn}(AP(\Phi_{L-ReLU}(Conv1D(\textit{\textbf{Z}}_{S}, (3, 1))))))
\end{equation}

Finally, the latent representation of $\textit{\textbf{Z}}_{fusion}$ will be fed into fully connected layers. The final output layer is activated by the softmax function, $\Phi_{softmax}(\cdot)$. Hence the final output can be calculated by:
\begin{equation}\label{eq:fc}
     \textbf{Output} = \Phi_{softmax}(\textit{\textbf{W$\mathcal{'}$}}\Phi_{dp}(\Phi_{ReLU}(\textit{\textbf{W}}(\Gamma(\textit{\textbf{Z}}_{fusion})) + b)) +b')
 \end{equation}
 where the $\Gamma(\cdot)$ is the squeeze operation, $\textit{\textbf{W}}$ and \textit{\textbf{W$\mathcal{'}$}} are the trainable weight matrix, $b$ and ${b}'$ are the bias terms.

The proposed TSception is summarised in \textbf{Algorithm~\ref{alg:TSception}}. The structure of the proposed TSception is shown in TABLE~\ref{Tab:TS_structure}
\begin{algorithm}
\SetAlgoLined
\SetKwBlock{DoParallel}{do in parallel}{end}
\SetKwBlock{DoSequential}{do in sequential}{end}
\KwIn{ EEG data $\textit{\textbf{X}}^{n} \in \mathbb{R}^{c \times l}$; ground truth label $y$;}

\KwOutput{$\widehat{y}$, the prediction of TSception}
Initialization\;
\DoSequential{
        \# \textit{get the output of the dynamic temporal layer} \\
        \For{$i\gets1$ \KwTo $3$}{
            get $i$-th temporal kernel size by \textbf{Eq.~\ref{eq:size_t}}\;
            get $\textbf{z}_{temporal}^{i}$ by \textbf{Eq.~\ref{eq:out_T}} using $\textit{\textbf{X}}^{n}$ as input\;
        }
        get $\textbf{Z}_{T}$ by \textbf{Eq.~\ref{eq:final_out_T}}\;
        \# \textit{get the output of the asymmetric spatial layer} \\
        \For{$j\gets0$ \KwTo $1$}{
            get $j$-th temporal kernel size by \textbf{Eq.~\ref{eq:size_S}}\;
            get $\textbf{z}_{spatial}^{j}$ by \textbf{Eq.~\ref{eq:out_T}} using $\textbf{Z}_{T}$ as input\;
        }
        get $\textbf{Z}_{S}$ by \textbf{Eq.~\ref{eq:final_out_S}}\;
        \# \textit{get the output of the high-level fusion layer} \\
        get $\textbf{Z}_{fusion}$ by \textbf{Eq.~\ref{eq:final_out_G_H}}\;
    }
 get $\widehat{y}$ using \textbf{Eq.~\ref{eq:fc}}\;
 \KwReturn{$\widehat{y}$}
 \caption{TSception}
 \label{alg:TSception}
\end{algorithm}


\section{Experiments}
\subsection{Datasets}
To evaluate the proposed TSception, we conducted several experiments on two publicly available benchmark datasets, a Database for Emotion Analysis using Physiological signals (DEAP)\footnote{http://www.eecs.qmul.ac.uk/mmv/datasets/deap/index.html} \cite{5871728}, and a multimodal database for affect recognition and implicit tagging (MAHNOB-HCI)\footnote{https://mahnob-db.eu/hci-tagging/} \cite{soleymani2011multimodal}. Table~\ref{Tab:dataset} summarizes the related information of the two datasets used in our experiments. Arousal and valence dimensions on both datasets were utilized as reported in \cite{9141493}.

DEAP is a multi-modal human affective states dataset, including EEG, facial expressions, and galvanic skin response (GSR). There are 32 subjects watching music video clips while their EEG, facial expression, and GSR are recorded. Each of the subjects participates in 40 trials in total. The duration of each trial is 1 minute with a 3 seconds pre-trial baseline. After each trial, the subject will be given a questionnaire to provide their own emotional state in arousal, valence, dominance, and liking with each dimension having 9 discrete levels. The EEG is collected using 32 channels device, with the sampling rate being 512Hz.

MAHNOB-HCI \cite{soleymani2011multimodal} is another multi-modal dataset similar to the DEAP dataset. There are 30 subjects watching movie clips while their facial expression, audio signals, eye gaze data, EEG signal, and other physiological signals are recorded. Note that Subject 12, 15, and 26 failed to finish the data collection, therefore, the remaining 27 out of 30 subjects were used in this work. The movie clips are between 35 and 117 seconds long. The EEG signals are acquired from $32$ electrodes on the $10$-$20$ international system. The sampling frequency is $256$ Hz. For each trial, four integers ranged from $1$ to $9$ and self-reported by the subjects are used to label the valence, arousal, dominance, and emotional keywords, respectively.

\begin{table}[htp]
\caption{Summary of related information of the datasets used in the experiments}
\begin{center}
\begin{tabularx}{0.45\textwidth}{
  >{\arraybackslash}X
  >{\arraybackslash}X 
  >{\arraybackslash}X
  }
 \hline
\textbf{Factor}&\textbf{DEAP} &\textbf{MAHNOB-HCI}\\
\hline
Subjects& 32 & 27 \\
Stimuli& Music videos& Emotional videos\\
Trials/subject& 40 & 20\\
Trial duration& 1 min & 35-117s\\
EEG channels& 32 & 32\\
Sampling rate& 512Hz & 256Hz\\
Label& V/A& V/A \\
\hline
\end{tabularx}
\begin{tablenotes}
      \small
      V: valence; A: arousal\\
      \item 
    \end{tablenotes}
\label{Tab:dataset}
\end{center}
\end{table}

\begin{table*}[htp]
\caption{Structure of the proposed TSception}
\begin{center}
\begin{tabularx}{1\textwidth}{sslmm}
\hline
Model structure& &Layers & Input& Output\\
\hline
Block1& 3 branches&Conv2d, LK-ReLU, AP(\com{(1,8)}) & (-1, 1, 28, 512) & (-1, 15, 28, 56) \\
&(in parallel) & Kernel=\com{15@(1, 64)}& &\\ \cline{3-5}
& &Conv2d, LK-ReLU, AP(\com{(1,8)}) &(-1, 1, 28, 512) & (-1, 15, 28, 60)\\
& &Kernel=\com{15@(1, 32)} & &\\ \cline{3-5}
& &Conv2d, LK-ReLU, AP(\com{(1,8)}) &(-1, 1, 28, 512) & (-1, 15, 28, 62)\\
& &Kernel=\com{15@(1, 16)} & &\\ \cline{2-5}
& &Concatenate, BN& &(-1, 15, 28, 178)\\
\hline
Block2& 2 branches&Conv2d, LK-ReLU, AP(\com{(1,2)}) &(-1, 15, 28, 178) & (-1, 15, 1, 89)\\
& (in parallel)& Kernel=\com{15@(28, 1)}& &\\ \cline{3-5}
& &Conv2d, LK-ReLU, AP(\com{(1,2)}) &(-1, 15, 28, 178) & (-1, 15, 2, 89)\\
& &Kernel=\com{15@(14, 1)}, Stride=\com{(14, 1)} & &\\ \cline{2-5}
& &Concatenate, BN& & (-1, 15, 3, 89)\\
\hline
Block3& &Conv2d, LK-ReLU, AP(\com{(1,4)}), BN, GAP &(-1, 15, 3, 89) &(-1, 15, 1)\\
& & Kernel=\com{15@(3, 1)}& &\\ \cline{3-5}
& & Flatten&(-1, 15, 1) & (-1, 15,)\\
\hline
Fully& & Linear(\com{32}), ReLU& (-1, 15,) & (-1, 32,)\\
connected& & dropout(\com{0.5})&(-1, 32,) & (-1, 32,)\\
layers& & Linear(\com{2})&(-1, 32,) & (-1, 2,)\\
& & softmax& (-1, 2,)& (-1, 2,)\\
\hline
\end{tabularx}
\begin{tablenotes}
      \small
      \item LK-ReLU is the Leaky-ReLU activation function. AP is the average pooling operation. BN stands for batch normalization. GAP is the global average pooling. '-1' in the tensor size stands for the number of samples within one mini-batch. The strides of CNNs are (1, 1) if not specified, and the one for pooling layers is the same as the pooling step.
    \end{tablenotes}
\label{Tab:TS_structure}
\end{center}
\end{table*}
\subsection{Pre-processing}
For DEAP, the 3 seconds pre-trial baseline was removed for each trial. Then the data was down-sampled from 512Hz to 128Hz, after which the electrooculogram (EOG) was removed with a blind
source separation method as \cite{5871728}. To remove the low and high-frequency noise, a band-pass filter from 4.0-45Hz was applied to the original EEG as \cite{5871728}. Finally, the EEG channels were averaged to the common reference. The class label for each dimension is from 1 to 9, hence 5 was selected as a threshold to project the 9 discrete values into low and high classes in each dimension as \cite{5871728}\cite{9141493}. In line with \cite{9141493}, only arousal and valence dimensions are used in this study. The deep neural networks have a higher number of trainable parameters hence to optimally learn emotion state representations in EEG a large number of labelled data samples are required. However, as listed in Table~\ref{Tab:dataset}, the number of trials is very small in the selected datasets. To overcome this challenge, a data augmentation step by splitting each trial into smaller non-overlapping 4s segments was applied. The segments were then used to train the deep neural network.

For MAHNOB-HCI, the pre-processing was much the same as that for the DEAP dataset except for the following. First, the 30 seconds pre-trial and post-trial baselines were removed for each trial, so that the remaining corresponds to the event of emotion elicitation \cite{soleymani2011multimodal}. Second, to remove the low-high frequency noise, a band-pass filter from $0.3$-$45$Hz was applied to the original EEG as \cite{9206750}. Note that the delta band $0.3$-$4$Hz is included since it also contributes to an individual's affective state \cite{knyazev2012eeg, daly2019electroencephalography}.

\subsection{Performance Evaluation Metrics}
The first type of metric is accuracy. It is one of the most commonly used evaluation metrics in classification problems \cite{9206750}. It is the ratio of the correctly predicted samples and the total number of the samples. For binary classification problems, the accuracy can also be defined as:
\begin{equation}\label{eq:acc}
     Accuracy = \frac{TP+TN}{TP+FP+TN+FN}
 \end{equation}
 where $TP$ is the true positive, $TN$ is the true negative, and $FP$ is the false positive, and $FN$ is the false negative.

Accuracy can measure how precise the prediction is for the class-balanced dataset. However, after the pre-processing of the labels mentioned in the pre-processing section, the labels become imbalanced. To better evaluate the performance of a classifier on class-imbalanced datasets, the F1 score is added as \cite{5871728}\cite{LIANG2019257}. It combines the precision and recall of the classifier, and it is defined as the harmonic mean of the classifier’s precision and recall. F1 is defined by:
\begin{equation}\label{eq:f1}
     F1 = 2\times \frac{Precision\times Recall}{Precision+Recall}=\frac{TP}{TP+\frac{1}{2}(FP+FN)}
 \end{equation}
 where $TP$ is the true positive, $TN$ is the true negative, and $FP$ is the false positive, and $FN$ is the false negative.

\subsection{Experiment Settings}
There are two types of experiment settings in this paper: I) trial-wise 10-fold cross-validation and II) leave-one-trial-out cross-validation. Each of them is introduced in the following paragraphs.

In the first experiment setting, we split each trial into 4's non-overlapping segments, also know as cropped experiments \cite{doi:10.1002/hbm.23730}, and a trial-wise 10-fold cross-validation is utilized for each subject to prevent potential data leakage issues. The reason for doing cropped experiments is that the predictions of shorter segments are preferred than the trial-wise predictions that are evaluated in \cite{5871728}\cite{LIANG2019257}\cite{9141493} for an efficient real-time BCI system. Besides, a decoding model with a good generalization capability is needed for the real-world situation where the testing data is unseen to the model. In each trial, the subject was asked to watch or hear a certain stimulus that is supposed to evoke a certain type of emotion. Because emotion is one of the continuous cognitive processes in the brain, the data segments within a single trial are highly correlated. Hence, randomly shuffle the segments among trials before the training-testing split of the data could make the adjacent segments be in training and testing data, which will give high classification results. But the accuracy will drop when the highly correlated segments are never seen by the model in the real-world situation. To get the more generalized evaluation, the 10 folds are split among trials, which will make sure the adjacent segments in one trial will not appear in both training and testing data. In each step of 10-fold cross-validation, one fold is selected as testing data, the rest 9 folds are utilized as training data. Among the 9 training folds, the data is randomly divided into 80\% training data and 20\% validation data. During the training process, we train the network on training data for 500 epochs and evaluate the network on validation data in each epoch. The model with the highest accuracy on validation data among those 500 epochs is saved and tested on the testing data. The above process is repeated 10 times for each subject till each fold has been the testing fold once. In each fold, the test data remains completely unseen in all stages of training and validation. The mean accuracy and F1 score of all subjects are reported as the final results. 

\begin{figure*}[htp]
    \centering
    \includegraphics[width= 18cm]{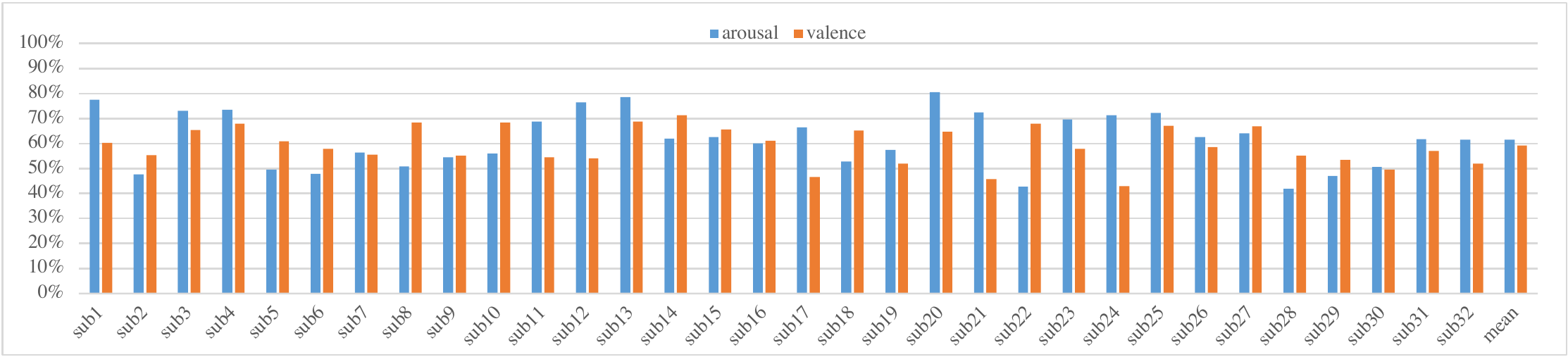}
    \caption{Mean accuracy of each subject for arousal and valence on DEAP using TSception.}
    \label{fig:acc-DEAP}
\end{figure*}
\begin{figure*}[htp]
    \centering
    \includegraphics[width= 18cm]{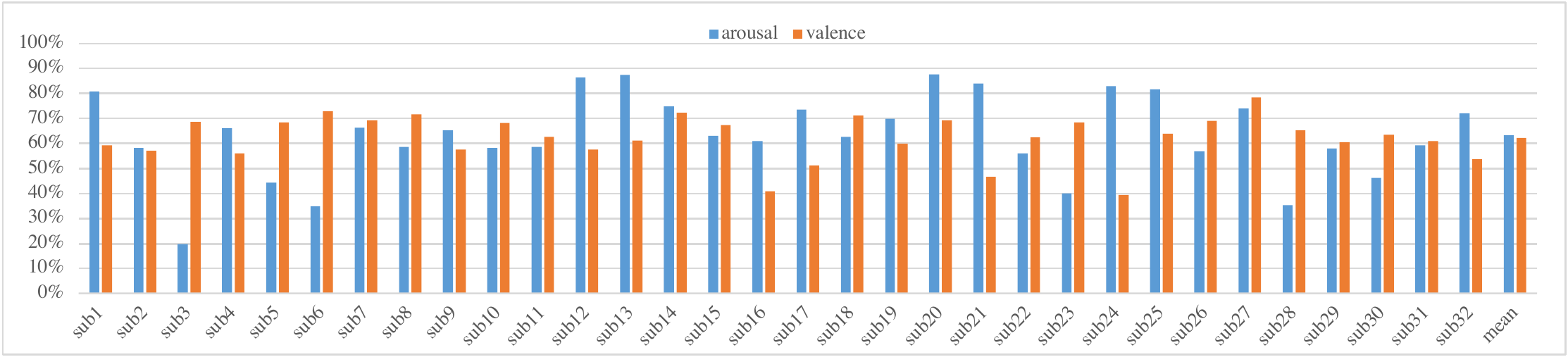}
    \caption{Mean F1 score of each subject for arousal and valence on DEAP using TSception.}
    \label{fig:f1-DEAP}
\end{figure*}

In the second experiment setting, a leave-one-trial-out cross-validation is adopted for each subject to further compare our methods with the recently proposed methods in \cite{9141493} and \cite{LIANG2019257}. In each cross-validation step, one trial is selected as testing data and the rest are selected as training data. For each step of the leave-one-trial-out cross-validation, the training data is also split into 80\% training and 20\% validation data. The process is repeated till every trial is selected as testing data once for each subject. The average accuracy and F1 score of all subjects are reported as the final evaluation criterion as \cite{LIANG2019257}. The features extracted from the entire trial's data are utilized as one input sample to the classifier in \cite{5871728}\cite{LIANG2019257}\cite{9141493}. To compare our deep learning methods trained from segmented EEG data with those papers, a voting mechanism is utilized for the segment predictions in each testing trial as:
\begin{equation}\label{eq:voting}
     \widehat{y}_{t} = \left\{\begin{matrix}
0 & n_{\widehat{y}=0} > n_{\widehat{y}=1}\\ 
1 & n_{\widehat{y}=1} \geq n_{\widehat{y}=0}
\end{matrix}\right.
 \end{equation}
 where $\widehat{y}_{t}$ is the prediction of one testing trial, $n$ is the number of the predictions of the segments in each trial under each condition indicated in the sub-script.

\subsection{Implementation Details}
The code is implemented using the PyTorch library, the source code can be found via this link\footnote{https://github.com/yi-ding-cs/TSception}. 

The ratio coefficients of T kernel length are [0.5, 0.25, 0.125] for DEAP. The sampling rate of the data in DEAP is 128Hz, hence, the temporal kernel lengths are 64, 32, and 16 according to Eq.~\ref{eq:size_t}. When training TSception on MAHNOB-HCI, we found that using [0.25, 0.125, 0.0625] as the ratio coefficients achieved higher mean accuracy on validation set. The sampling rate of MAHNOB-HCI is 256Hz, which gives the temporal kernel lengths of 64, 32, and 16 as well. The number of temporal and spatial kernels in dynamic temporal, asymmetric spatial, and high-level fusion layers is equal to 15. The number of hidden nodes in the first fully connected layer is chosen as 32. For model training, the maximum training epoch is 500. The batch size on the DEAP dataset is set as 64 which will be reduced to 32 on the MAHNOB-HCI dataset because the trials in MAHNOB-HCI are half of the ones in DEAP. All the other hyper-parameters (including the structure hyper-parameters as well as the training hyper-parameters), except batch size, are the same for DEAP and MAHNOB-HCI to test the generalization ability of TSception. The hyper-parameters are the same for all the subjects. Adam optimizer is utilized to optimize the training process with the initial learning rate being 1e-3. Cross-entropy loss is selected as the loss function to guide the training process. For more details, please refer to the open-access GitHub repository for TSception.

\section{Results and Analysis}
In this section, we first report and statistically compare the results in terms of accuracy and F1 score for ours against the state-of-the-art methods. The ablation studies are then presented to reveal the contribution of each component in TSception. Finally, saliency maps are presented to visualize how the brain areas contribute to the arousal and valence dimensions.

\begin{table*}[htp]
\caption{Trial-wise 10-fold cross-validation classification results of SVM, KNN, EEGNet, ShallowConvNet, DeepConvNet, and TSception on DEAP}
\begin{center}
\begin{tabularx}{1\textwidth} {
  >{\arraybackslash}X
  >{\arraybackslash}X 
  >{\arraybackslash}X
  >{\arraybackslash}X 
  >{\arraybackslash}X
  >{\arraybackslash}X 
  >{\arraybackslash}X
  >{\arraybackslash}X 
  >{\arraybackslash}X
  >{\arraybackslash}X
  }
 \hline
&\multicolumn{4}{c}{\textbf{Arousal}}&\multicolumn{4}{c}{\textbf{Valence}}& \\  \cline{2-9}
\textbf{Method}&\textbf{\textit{ACC}}&std &\textbf{\textit{F1}}&std&\textbf{\textit{ACC}}&std &\textbf{\textit{F1}}&std&\textbf{Parameters}\\
\hline
SVM& 60.37\% &12.25\%& 57.33\% &26.61\% & 55.19\% **&6.97\%& 57.87\%** &11.36\%& NA\\
KNN& 59.48\% &12.34\%& 57.49\% &24.96\%& 53.03\% **&9.14\%& 55.12\%**&16.27\%&NA\\
EEGNet&58.29\%& 8.60\% & 60.60\% &15.20\%& 54.56\% **&8.14\%& 57.61\%**&10.42\%&2,162\\
SCN& 61.19\%&10.28\%& 61.19\% &20.08\%& 59.42\% &8.30\%& 62.26\% &11.49\%&48,162\\
DCN&61.03\%& 8.58\% & 62.58\% &17.40\%& \textbf{59.92\%} &7.82\%& 62.04\% &10.23\%&151,252\\

\hline
\textbf{TSception}&\textbf{61.57\%}& 11.04\% & \textbf{63.24\%} &16.60\%& \textbf{59.14\%}&7.60\%& \textbf{62.33\%} &9.03\%&\textbf{12,563}\\
\hline
\end{tabularx}
\begin{tablenotes}
      \small
      \item $p$-value between the method and TSception: * indicating $(p < 0.05)$, ** indicating $(p<0.01)$, *** indicating $(p < 0.001)$.
      \item SCN: ShallowConvNet
      \item DCN: DeepConvNet
    \end{tablenotes}
\label{Tab:result_DEAP_10_fold}
\end{center}
\end{table*}

\begin{table*}[htp]
\caption{Trial-wise 10-fold cross-validation classification results of SVM, KNN, EEGNet, ShallowConvNet, DeepConvNet, TSception on MAHNOB-HCI}
\begin{center}
\begin{tabularx}{1\textwidth} {
  >{\arraybackslash}X
  >{\arraybackslash}X 
  >{\arraybackslash}X
  >{\arraybackslash}X 
  >{\arraybackslash}X
  >{\arraybackslash}X 
  >{\arraybackslash}X
  >{\arraybackslash}X 
  >{\arraybackslash}X
  >{\arraybackslash}X
  }
 \hline
&\multicolumn{4}{c}{\textbf{Arousal}}&\multicolumn{4}{c}{\textbf{Valence}}& \\  \cline{2-9}
\textbf{Method}&\textbf{\textit{ACC}}&std &\textbf{\textit{F1}}&std&\textbf{\textit{ACC}}&std &\textbf{\textit{F1}}&std&\textbf{Parameters}\\
\hline
SVM& 58.25\% *&14.09\%& \textbf{33.40\%} &21.87\% & 58.44\% *&9.39\%& 40.27\% &12.73\%& NA\\
KNN& \textbf{60.95\%} &17.11\%& 28.65\% &24.94\%& 60.32\% &12.28\%& 28.62\%*** &20.26\%&NA\\
EEGNet&59.98\%& 16.16\% & 30.47\% &23.68\%& 56.43\% ***&11.12\%& 33.98\%** &15.27\%&2,674\\
SCN& 59.85\% &16.02\%& 30.60\% &22.20\%& 59.57\%&11.25\%& 36.41\%**&15.27\%&50,882\\
DCN&57.29\%& 15.69\% & 32.37\% &23.69\%& 60.29\% &12.38\%& 36.09\%*&17.64\%&153,652\\

\hline
\textbf{TSception}&\textbf{60.61\%}& 14.88\% & \textbf{33.06\%} &23.35\%& \textbf{61.27\%}&10.05\%& \textbf{40.66\%} &16.52\%&\textbf{12,563}\\
\hline
\end{tabularx}
\begin{tablenotes}
      \small
      \item $p$-value between the method and TSception: * indicating $(p < 0.05)$, ** indicating $(p<0.01)$, *** indicating $(p < 0.001)$. 
      \item SCN: ShallowConvNet
      \item DCN: DeepConvNet
    \end{tablenotes}
\label{Tab:result_MAHNOB}
\end{center}
\end{table*}

\begin{table}[htp]
\caption{Compare with the results reported in the existing literatures using leave-one-trial-out cross-validation on DEAP}
\begin{center}
\begin{tabularx}{0.45\textwidth} {bssss}
 \hline
&\multicolumn{2}{c}{\textbf{Arousal}}&\multicolumn{2}{c}{\textbf{Valence}}\\  \cline{2-5}
\textbf{Method}&\textbf{\textit{ACC}}&\textbf{\textit{F1}}&\textbf{\textit{ACC}}&\textbf{\textit{F1}}\\
\hline
SVM\cite{5871728}& 62.00\% &58.30\%& 57.60\% &56.30\%\\
UL\cite{LIANG2019257}& 62.34\% &60.44\%& 56.25\% &61.25\%\\
CSP\cite{9141493}&58.26\%& - & 57.59\% &-\\
FBCSP\cite{9141493}& 59.13\% &-& 59.19\% &-\\
FgMDM\cite{9141493}& 60.04\% &-& 58.87\% &-\\
TSC\cite{9141493}& 60.04\% &-& 59.47\% &-\\
FBFgMDM\cite{9141493}&60.30\%&-& 61.01\% &-\\
FBTSC\cite{9141493}&60.60\% &-& 61.09\% &-\\
\hline
\textbf{TSception(ours)}&\textbf{63.75\%}& \textbf{63.35\%} & \textbf{62.27\%} &\textbf{65.37\%}\\
\hline
\end{tabularx}
\label{Tab:result_DEAP_LOTO}
\end{center}
\end{table}
\subsection{Statistical Analysis}
The experiment results: include I) the per-subject accuracy and F1 score on DEAP dataset (see Fig.~\ref{fig:acc-DEAP} and Fig.~\ref{fig:f1-DEAP}), II) the overall accuracy and F1 score on DEAP dataset (see Table~\ref{Tab:result_DEAP_10_fold}) MAHNOB-HCI data set (see Table~\ref{Tab:result_MAHNOB}), and III) the comparison against the results from existing literatures (see Table~\ref{Tab:result_DEAP_LOTO}). To conduct statistical analysis, a two-tailed Wilcoxon Signed-Rank Test is utilized. Compared to accuracy, F1 score is a more reliable metric to quantify the performance of classification methods when a dataset has imbalanced classes. Based on the results we have the following observation and analysis. 

On DEAP, deep learning methods generally outperform non-deep learning methods, whereas, on MAHNOB-HCI, the SVM outperforms EEGNet, ShallowConvNet, and DeepConvNet, and is comparable to our method. On DEAP, ours outperforms SVM and KNN on all the experiments. Ours has a 1.2\% higher accuracy and a 5.91\% higher F1 score than SVM for the arousal dimension. For valence, ours outperform SVM with the improvement in accuracy and F1 score being 3.95\% ($p<0.05$) and 4.46\% ($p<0.05$) respectively. Compared with KNN, ours achieves 2.09\%/5.91\% higher accuracy/F1 score for arousal and 3.95\% ($p<0.05$)/7.21\% ($p<0.05$) higher accuracy/F1 score for valence. On MAHNOB-HCI, ours achieves the best accuracy and F1 score on valence, while the best accuracy on arousal is achieved by KNN classifier and the best F1 score is the one using SVM. But TSception still achieves a 4.41\% higher F1 score than KNN ($p=0.07186$) and a 2.36\% higher accuracy than SVM ($p<0.05$) for arousal dimension. Although the other three deep learning methods have higher classification results the SVM and KNN on DEAP, they have lower accuracy or F1 score than those non-deep learning methods. SVM defeats EEGNet, ShallowConvNet, and DeepConvNet on F1 scores for both arousal and valence dimensions. And KNN achieves better accuracies than those three deep learning methods for both arousal and valence. It further suggests that except for ours, most of the deep learning methods demonstrate less cross-dataset generality comparing to SVM and KNN. 

Among the four deep learning methods, ours achieves the highest accuracy and F1 score in most of the experiments on DEAP. In particular, our TSception (12,563) has only a quarter and one-tenth of the parameters compared to ShallowConvNet (48,162) and DeepConvNet (151,252), respectively. TSception achieves the highest accuracy and F1 score on arousal as well as the highest F1 score on valence, with the accuracy being 61.57\% for arousal, 59.14\% for valence, and the F1 score being 63.24\% for arousal, 62.33\% for valence respectively. DeepConvNet achieves second place compared with the other methods (accuracy: 61.03\% for arousal, 59.92\% for valence, F1: 62.58\% for arousal, 62.04\% for valence). ShallowConvNet gets the third-highest results among all the compared methods, achieving 61.19\%/59.42\% for arousal/valence in terms of accuracy, and 61.19\%/62.26\% for arousal/valence in terms of F1 score. The accuracy of TSception for arousal is 3.28\% higher than EEGNet ($p=0.05118$), the one for valence has a 4.58\% improvement over EEGNet ($p<0.01$). For the F1 score, TSception has a 2.64\% higher F1 score than EEGNet for arousal ($p=0.05614$), and a 5.72\% higher F1 score for valence ($p<0.01$).

On MAHNOB-HCI, ours achieves the highest accuracy and F1 score among four deep learning methods. For accuracy metrics, TSception is 3.32\% ($p=0.0536$) and 0.98\% ($p=0.35238)$ higher than DeepConvNet for arousal and valence respectively. Compared with ShallowConvNet, TSception has higher accuracies for arousal (0.76\%) and valence (1.7\%) with the p-value being 0.48392 and 0.12356. TSception is 0.63\% ($p= 0.87288$) and 4.84\% ($p<0.001$) higher than EEGNet for arousal and valence in terms of accuracy. For F1 scores, TSception achieves much higher results than the other three deep learning methods. Especially for valence dimension, TSception achieves 4.57\% ($p<0.05$), 4.25\% ($p<0.01$), and 6.68\% ($p<0.01$) higher F1 score than DeepConvNet, ShallowConvNet, and EEGNet.

Interestingly, we also notice that the difficulty to predict the two emotional dimensions are not consistent for the two datasets. Considering the trade-off of accuracy and F1 score, we find that the valence is harder to predict for DEAP while the arousal is harder to predict for MAHNOB-HCI. 

Our method outperforms the results reported in the existing literatures \cite{5871728}\cite{LIANG2019257}\cite{9141493} as well. According to Table~\ref{Tab:result_DEAP_LOTO}, ours achieves the best accuracies for both arousal and valence dimensions. TSception has 3.15\% and 1.18\% improvements over FBTSC \cite{9141493} on accuracy for arousal and valence. Compared with UL \cite{LIANG2019257}, our method beats it by 1.41\% for arousal and 6.02\% for valence in terms of accuracy. The accuracies of ours for arousal and valence are 1.75\% and 4.67\% higher than the ones of SVM reported in \cite{5871728}. For F1 scores, TSception has 5.05\% and 2.91\% higher than SVM \cite{5871728} and UL \cite{LIANG2019257} for arousal and 9.07\% and 4.12\% higher for valence, indicating the effectiveness of the proposed method. 

According to the extensive comparison against a variety of methods, the proposed method manifests promising performance on the arousal-valence prediction task, with a decent 
extent of generality.

\begin{table}[htp]
\caption{Ablation study results of removing functional layers in TSception using DEAP}
\begin{center}
\begin{tabularx}{0.45\textwidth} {
  >{\arraybackslash}X
  >{\arraybackslash}X 
  >{\arraybackslash}X
  >{\arraybackslash}X 
  >{\arraybackslash}X
  }
 \hline
&\multicolumn{2}{c}{\textbf{Arousal}}&\multicolumn{2}{c}{\textbf{Valence}}\\  \cline{2-5}
\textbf{Method}&\textbf{\textit{ACC}} &\textbf{\textit{F1}}&\textbf{\textit{ACC}} &\textbf{\textit{F1}}\\
\hline
w/o  T& 60.45\%& 61.29\% & 58.62\% & 61.47\% \\
w/o  S& 60.07\%& 61.41\% & 56.90\%& 60.74\% \\
w/o  F&60.03\%& 60.85\% & 58.21\% & 61.14\%\\
\hline
\textbf{TSception}&\textbf{61.57\%}& \textbf{63.25\%} & \textbf{59.14\%} & \textbf{62.33\%} \\
\hline
\end{tabularx}
\begin{tablenotes}
      \small
      T: Dynamic temporal layer; S: Asymmetric spatial layer; F: High-level fusion layer.\\
      w/o: Without the component.\\
    \end{tablenotes}
\label{Tab:result_ablation}
\end{center}
\end{table}

\subsection{Ablation Study}

The proposed method TSception has a dynamic temporal layer, asymmetric spatial layer, and high-level fusion layer three functional parts. The combination of those three parts leads to the success of classification tasks. Ablation studies are conducted to further understand which part contributes more to the improvement of classification results. The classification results after removing each of the dynamic temporal layer, asymmetric spatial layer, and high-level fusion layer from the TSception are reported. DEAP dataset is used for the ablation study since the overall performance is higher than MAHNOB-HCI. The results of the ablation study are shown in Table ~\ref{Tab:result_ablation}. 

All of the accuracies and F1 scores drop after removing any of the three types of layers, indicating all components contribute to the improvement of classification results. Overall, the most significant drops of accuracy for three dimensions are observed when the asymmetric spatial layer is removed from TSception with the decrements being 1.5\%/1.84\% on accuracies for arousal/valence and 1.84\%/1.59\% on F1 scores for arousal/valence. This demonstrated that the asymmetric spatial layer contributes more than the other two layers, especially for the valence dimension, the drop is the largest in the ablation study. The high-level fusion layer contributes more to arousal because the accuracy drops by 1.54\%, and the F1 score drops by 2.40\% for arousal while the drops of accuracy and F1 score for valence are smaller (0.93\% on accuracy and 1.19\% on F1 score) after removing the high-level fusion layer. The dynamic temporal layer contributes less than the others, with the drops of accuracy/F1 score being 1.12\%/1.96\% for arousal and 0.52\%/0.86\% for valence. 

The kernel-level ablation studies are further conducted to analyze the effects of two types of spatial kernels in the spatial asymmetric layer because it has more contribution than other layers. The weights and biases are set to zeros as \cite{Lawhern_2018} did to study the kernel-level effects. The results are shown in Table ~\ref{Tab:result_ablation_kernel}. 

Hemisphere kernels learn more discriminative representations than global kernels in TSception, according to the results in Table ~\ref{Tab:result_ablation_kernel}. The drops of classification results after removing the hemisphere kernels are all larger than the ones after removing the global kernels. After removing either type of the hemisphere and global kernels will downgrade the performance of TSception for both the arousal and valence dimensions. This indicates both types of spatial convolutions help to improve the performance of TSception. 

\begin{table}[htp]
\caption{Ablation study results of removing spatial convolutional kernels in TSception using DEAP}
\begin{center}
\begin{tabularx}{0.45\textwidth} {
  >{\arraybackslash}X
  >{\arraybackslash}X 
  >{\arraybackslash}X
  >{\arraybackslash}X 
  >{\arraybackslash}X
  }
 \hline
&\multicolumn{2}{c}{\textbf{Arousal}}&\multicolumn{2}{c}{\textbf{Valence}}\\  \cline{2-5}
\textbf{Method}&\textbf{\textit{ACC}} &\textbf{\textit{F1}}&\textbf{\textit{ACC}} &\textbf{\textit{F1}}\\
\hline
w/o  H& 55.86\%& 50.38\% & 51.15\% & 40.13\% \\
w/o  G& 57.21\%& 58.29\% &54.22\%& 57.75\% \\
\hline
\textbf{TSception}&\textbf{61.57\%}& \textbf{63.25\%} & \textbf{59.14\%} & \textbf{62.33\%} \\
\hline
\end{tabularx}
\begin{tablenotes}
      \small
      H: Hemisphere kernels; G: Global kernels.\\
      w/o: Without the component.\\
    \end{tablenotes}
\label{Tab:result_ablation_kernel}
\end{center}
\end{table}

\subsection{Interpretability}
\begin{figure}[htp]
\centering
\includegraphics[width=0.5\textwidth]{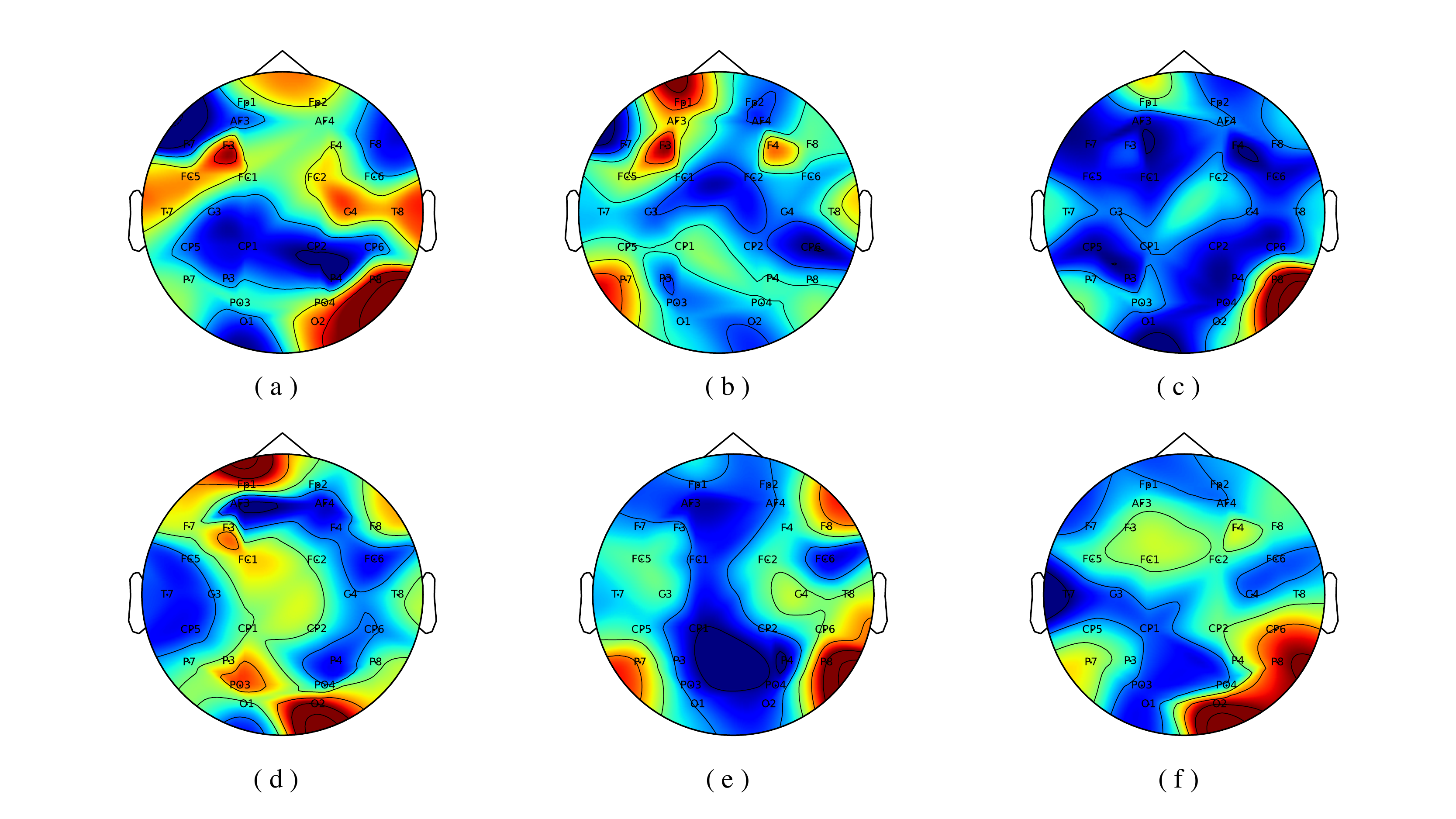}

\caption{Averaged saliency maps in DEAP dataset. The upper three saliency maps (a)-(c) are the averaged saliency maps for arousal dimension while the lower three (d)-(f) are for valence. The first column (a) and (d) are the mean saliency map of all the subjects. The second column (b) and (e) are the one of subjects who are top 10\% for F1 scores, The last column (c) and (f) are the average saliency map of the subjects whose F1 scores are in bottom 10\% for arousal. The mean saliency map is normalized between -1 and 1 for better visualization. F1 is chosen as the criterion because it can reflect how precise the predictions are by taking the imbalanced classes issue into consideration. The most informative region identified by the neural network is the frontal, temporal, parietal, and regions for high F1 score subjects.}
\label{fig:saliencymap_all}
\end{figure}

\begin{figure*}[htp]
\centering
\includegraphics[width=1\textwidth]{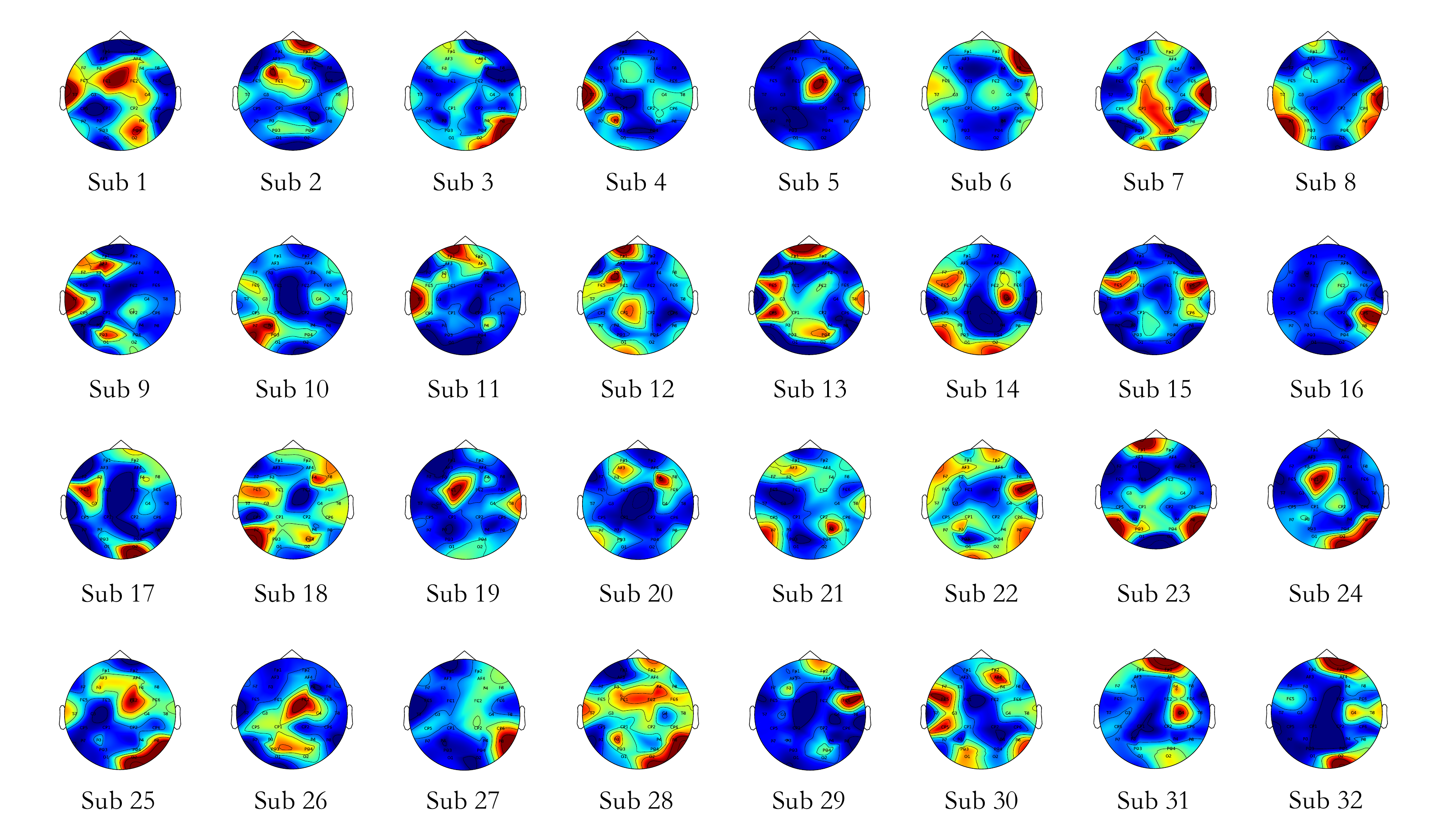}

\caption{Saliency maps of all 32 subjects for arousal in DEAP dataset. The saliency map is averaged along the time dimension to plot the topological map.}
\label{fig:saliency_map_subjects}
\end{figure*}

In this part, the saliency map \cite{simonyan2013deep} is utilized to visualize which parts of the data are more informative and contribute to classification performance. The saliency map is one of the most commonly used tools to intuitively show which regions of the input have the classification-related information. To better visualize the saliency map, the original saliency map is averaged along the time dimension to get the topological map of the EEG channels. The normalized saliency maps of different samples of each subject are averaged to get the mean saliency map of the subject for general visualization. The averaged saliency maps in the DEAP dataset are shown in Fig.~\ref{fig:saliencymap_all}. The mean saliency maps of individuals for arousal are also shown in Fig.~\ref{fig:saliency_map_subjects} to illustrate the differences across subjects. 

The pictures in Fig.~\ref{fig:saliencymap_all} are the saliency maps under different calculation settings. The upper three saliency maps, Fig.~\ref{fig:saliencymap_all}(a)-(c), are the averaged saliency maps for arousal dimension while the lower three, Fig.~\ref{fig:saliencymap_all}(d)-(f), are for valence. The first column, Fig.~\ref{fig:saliencymap_all}(a) and Fig.~\ref{fig:saliencymap_all}(d), are the mean saliency map of all the subjects. The second column, Fig.~\ref{fig:saliencymap_all}(b) and Fig.~\ref{fig:saliencymap_all}(e), are the one of subjects who are top 10\% for F1 scores, The last column, Fig.~\ref{fig:saliencymap_all}(c) and Fig.~\ref{fig:saliencymap_all}(f), are the average saliency map of the subjects whose F1 scores are in bottom 10\% for arousal. The mean saliency map is normalized between -1 and 1 for better visualization. We choose F1 as the selecting criterion for visualization because it can reflect how precise the predictions are when the classes are imbalanced. 

For arousal, the frontal, temporal, and right side of the parietal and occipital areas of the brain are more informative according to Fig.~\ref{fig:saliencymap_all}(a) and Fig.~\ref{fig:saliencymap_all}(b). The averaged saliency map of all the subjects, Fig.~\ref{fig:saliencymap_all}(a), shows the value of Fp2, F3, FC2, FC5, T7, T8, C4, P8, and O2 channels are higher than others. Comparing the saliency maps of the top(Fig.~\ref{fig:saliencymap_all}(b)) and bottom 10\% (Fig.~\ref{fig:saliencymap_all}(c)) F1 score subjects, we can see the frontal (Fp1, AF3, F3 and F4), temporal (T8) and parietal (P7) areas provide more information in Fig.~\ref{fig:saliencymap_all}(b), while the network mainly learns from parietal (P8) in Fig.~\ref{fig:saliencymap_all}(c). This indicates frontal, temporal, and parietal areas of the brain provide more emotion-related information. This is consistent with previous literatures \cite{gao2021novel}\cite{huang2012asymmetric}\cite{mickley2009effects}. Emotion arousal is mostly reflected in the frontal lobe, temporal lobe, and parietal lobe \cite{gao2021novel}. Pre-frontal and temporal asymmetry have close relations to arousal recognition \cite{huang2012asymmetric}. 

For valence, the frontal, temporal, and right side of the parietal and occipital areas of the brain are more informative according to Fig.~\ref{fig:saliencymap_all}(d) and Fig.~\ref{fig:saliencymap_all}(e). The same thing happens that the occipital (O1 and O2) activities provide less classification-related information than frontal (F8), temporal (T8), and parietal (P7 and P8) activities for the high F1 score subjects (Fig.~\ref{fig:saliencymap_all}(e)). According to previous studies, the asymmetry patterns in pre-frontal, parietal, and temporal regions are observed for valence recognition \cite{huang2012asymmetric}.

In general, the most informative region identified by the neural network is the frontal, temporal, parietal, and occipital regions while the occipital activities are less informative for the subjects with high F1 scores. This is consistent with previous works \cite{gao2021novel}\cite{huang2012asymmetric}\cite{li2019regional}\cite{shammi1999humour}\cite{kragel2018generalizable}, which indicates the network learns from the proper region. And for both arousal and valence, the occipital activities provide certain information. This may be because the stimuli used in DEAP are music videos. 

\section{Discussion and Conclusion}
Accurate emotion detection can benefit many healthcare applications including Cognitive Behavioural Therapy (CBT), Emotion Regulation Therapy (ERT)/Emotion-Focused Therapy (EFT) for emotion-related mental disorder treatment. Most of the previous works highly rely on the human extracted features, which requires heavy domain knowledge. Deep learning, especially the family of convolutional neural networks, has the auto feature-extracting ability. In this paper, we propose TSception, a multi-scale convolutional neural network, for EEG emotion recognition tasks. The parallel multi-scale temporal kernels whose lengths are related to the sampling rate of EEG are proposed in the temporal convolutional layer of TSception to enrich the learned temporal/frequency representations. To capture the emotional asymmetry patterns, we propose hemisphere kernels besides the global kernels in the asymmetric spatial layer. A high-level fusion layer is designed to further learn from the hemisphere/global representations of EEG and reduce the model size. 

To get the generalized evaluation of our method, we adopt the trial-wise cross-validation of cropped trials on two benchmark datasets. As mentioned in Section 4.4, if one randomly shuffles the samples among different trials before dividing the data into training and testing data in cropped experiments, he can get very high classification results that will drop when the highly correlated adjacent segments in one trial are not seen by the model \cite{LIANG2019257}\cite{9093122}. Hence, the trial-wise 10-fold cross-validation is utilized to make sure the highly correlated adjacent segments of each trial don't appear in both training and testing data. To further compare our methods with the ones in the existing literatures that also use generalized evaluation settings, a leave-one-trial-out cross-validation is conducted with a voting mechanism on each trial's segment predictions. As for evaluating metrics, we also follow \cite{LIANG2019257}, adding F1 score besides accuracy to get a better evaluation on imbalanced datasets. 

According to the results on two public datasets shown in Table~\ref{Tab:result_DEAP_10_fold}, Table~\ref{Tab:result_MAHNOB} and Table~\ref{Tab:result_DEAP_LOTO}, the proposed TSception achieves the highest classification results than those from the compared methods in most of the experiments. Particularly, TSception has 1/4 or 1/10 of the trainable parameters of its counterparts. Such efficiency and effectiveness may benefit the online usage of the neural network in real-world BCI applications. 

Extensive ablation studies and interpretability experiments suggested that all modules in TSception have positive contributions to the improvement of classification results and our method learns from the emotion-related information. According to Table~\ref{Tab:result_ablation}, we find the asymmetric spatial layer contributes most to the classification results. To make sure the neural network learns the emotion-related information instead of irrelevant features, saliency maps are acquired to visualize the most informative regions identified by the neural network itself. The mean saliency maps of all subjects in Fig.~\ref{fig:saliencymap_all}(a) and (d) show strong activation in the frontal, temporal, parietal, and occipital areas. However, the saliency maps of the subjects with high F1 scores in Fig.~\ref{fig:saliencymap_all}(b) and (e) only show strong activation in the frontal, temporal, and parietal areas, which is consistent with \cite{7946165}\cite{gao2021novel}\cite{huang2012asymmetric}\cite{li2019regional}\cite{shammi1999humour}\cite{kragel2018generalizable}\cite{dennis2010frontal}. A right hemisphere lateralization pattern is also observed in the averaged saliency map of the top 10\% subjects with high F1 scores (Fig.~\ref{fig:saliencymap_all}(e)) for valence, which indicates the right hemisphere is more informative for valence recognition. Neuroscience studies \cite{schwartz1975right}\cite{sheppard2020right} suggested that the right hemisphere has a special role in the emotional process in the brain. However, the right hemisphere lateralization is not present for high F1 subjects for arousal as shown in Fig.~\ref{fig:saliencymap_all}(b). This could be because the information provided in the frontal area is enough for the neural network to make the decision. Moreover, we find that the occipital activities also contribute to the inference process of the neural network for all the subjects, as shown in Fig.~\ref{fig:saliencymap_all}(a) and (d). A possible reason for high occipital activities is that music videos are used as stimuli in DEAP. However, the information provided by occipital activities is less useful for high F1 subjects (for both arousal and valence). This suggest occipital is not as informative as other brain regions, such as frontal and temporal regions, for emotion recognition.

To conclude, we propose a multi-scale convolutional neural network, named TSception, to capture temporal dynamics and spatial asymmetry for EEG emotion recognition. Using generalized cross-validation strategies, the proposed method and several baseline methods are evaluated on two publicly available benchmark datasets. The proposed method manifests promising performance on the arousal-valence prediction task, with a decent extent of generality. In the future, the generalization ability of TSception across subjects will be explored. The effect of segment length in cropped experiments on TSception should also be considered and studied.


%



\ifCLASSOPTIONcompsoc
  \section*{Acknowledgments}
  This work was partially supported by the RIE2020 AME Programmatic Fund, Singapore (No. A20G8b0102).

\ifCLASSOPTIONcaptionsoff
  \newpage
\fi



%

\bibliographystyle{./Bibliography/IEEEtran}
\bibliography{./Bibliography/mybib}




%








\end{document}